%% file: AI-HRI_2022.tex
\def\year{2022}\relax
\documentclass[letterpaper]{article} 
\usepackage{aaai22}  
\usepackage{times}  
\usepackage{helvet}  
\usepackage{courier}  
\usepackage[hyphens]{url}  
\usepackage{graphicx} 
\usepackage{natbib}  
\usepackage{caption} 
\DeclareCaptionStyle{ruled}{labelfont=normalfont,labelsep=colon,strut=off} 
\usepackage{algorithm}
\usepackage{algorithmic}
\usepackage{fancyvrb}
\usepackage{newfloat}
\usepackage{subcaption}
\usepackage{listings}
\floatstyle{ruled}
\newfloat{listing}{tb}{lst}{}
\floatname{listing}{Listing}
\title{A Socially Assistive Robot using Automated Planning\\in a Paediatric Clinical Setting}
\author {
    Alan Lindsay,\textsuperscript{\rm 1}
    Andr\'es Ram\'\i rez-Duque,\textsuperscript{\rm 2}
    Ronald P.A. Petrick,\textsuperscript{\rm 1}
    Mary Ellen Foster\textsuperscript{\rm 2}
}
\affiliations {
    \textsuperscript{\rm 1} Department of Computer Science, Heriot-Watt University, Edinburgh, UK \\
    \textsuperscript{\rm 2} School of Computing Science, University of Glasgow, Glasgow, UK \\
    \{Alan.Lindsay, R.Petrick\}@hw.ac.uk \hspace{1em} \{Andres.Ramirez-Duque, MaryEllen.Foster\}@glasgow.ac.uk
}
\copyrighttext{Presented at the AI-HRI Symposium at AAAI Fall Symposium Series (FSS) 2022}

\begin{document}

\maketitle

\begin{abstract}
\input{abstract}
\end{abstract}

\section{Introduction}

\input{introduction}

\begin{figure*}
\begin{subfigure}{0.62\textwidth}
\includegraphics[width=\textwidth]{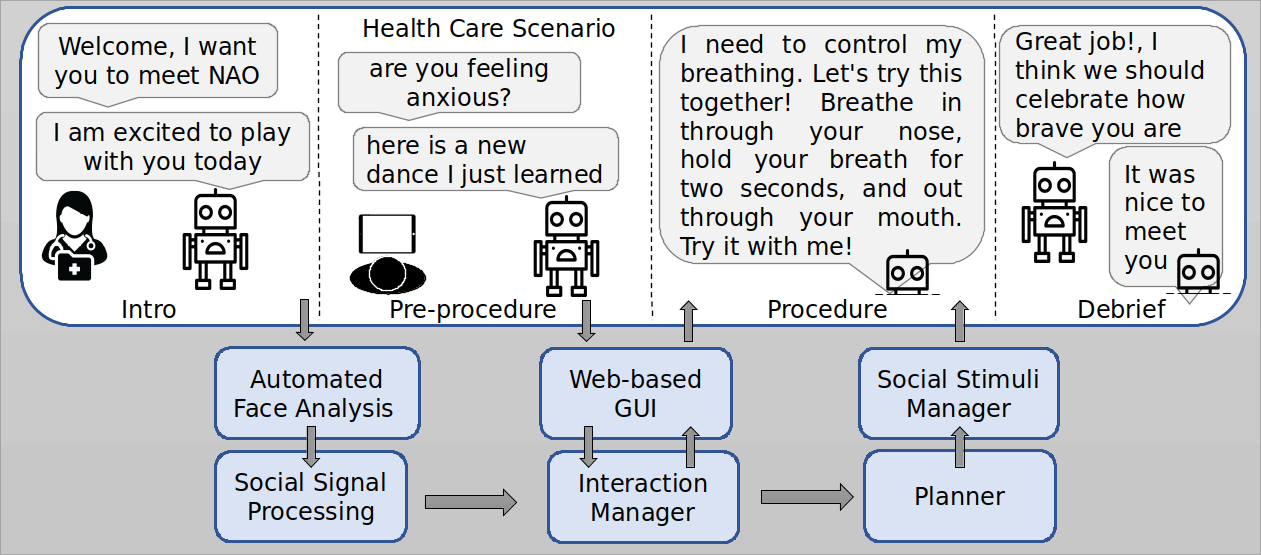} 
\caption{System architecture
including social signal processing, interaction management, and planning system. Robot utterances are drawn from previous studies \cite{ali2019lo63}.}
\label{fig:system}
\end{subfigure}
\hfill
\begin{subfigure}{0.35\textwidth}
\includegraphics[width=\textwidth]{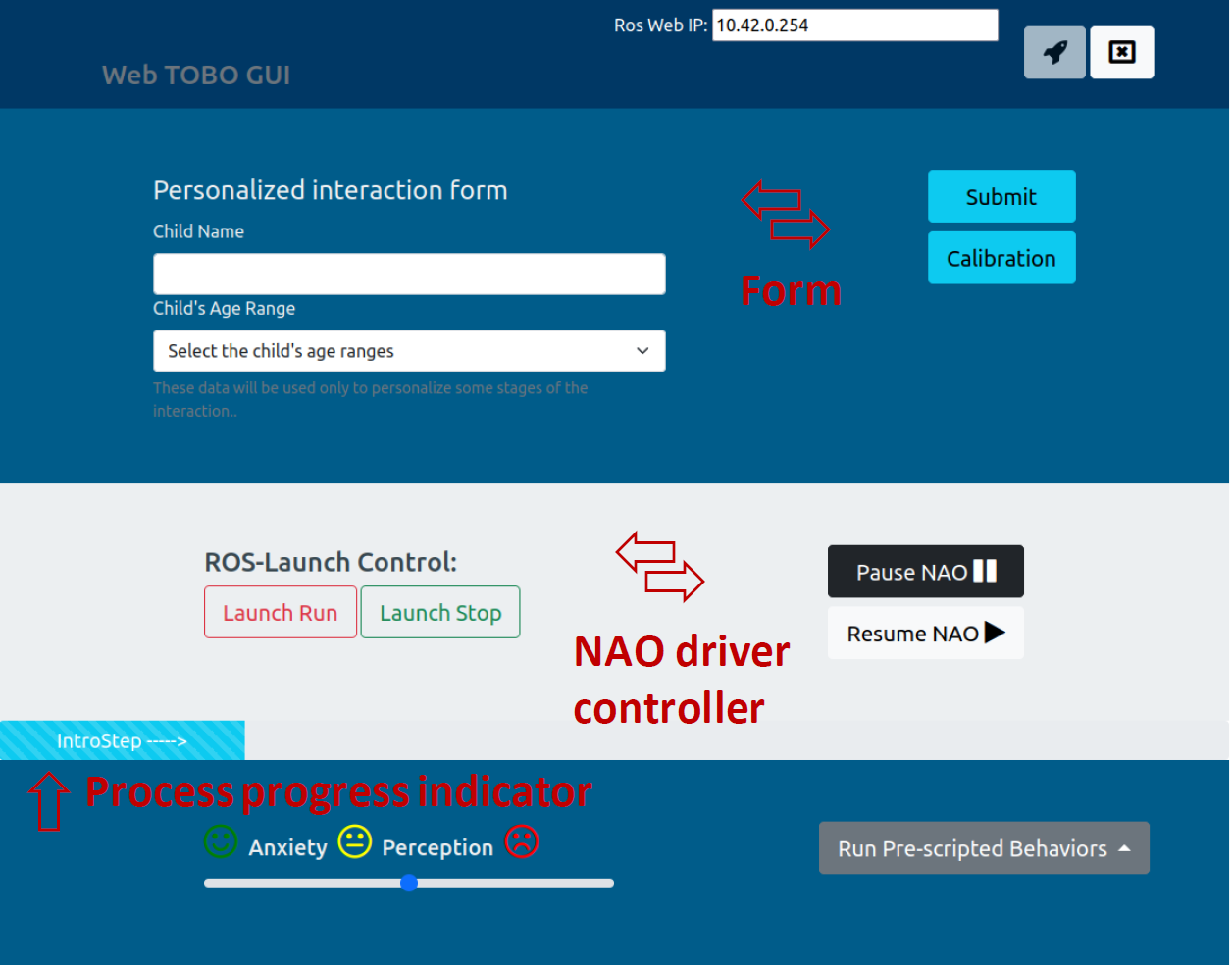}
\caption{GUI used to control the system and to confirm social signal processing hypotheses.}
\label{fig:gui}
\end{subfigure}

\caption{System architecture and GUI}
\end{figure*}

\section{Related Work}
\input{related_work}

\section{System Overview}

\input{system_overview}


\section{System Design}
The target setting is complex, dynamic, and requires sensitive interactions: it is therefore essential that the robot is robust and able to adapt to various (possibly adverse) situations, which can be difficult to predict in advance.

\input{procedure_modelling}

\input{world_knowledge}

\input{transition_modelling}

\input{robustness}

\input{conclusion}

\section{Acknowledgments}
The authors wish to acknowledge the SSHRC-UKRI Canada-UK Artificial Intelligence Initiative (UKRI grant ES/T01296/1) for financial support of this project. We would also like to acknowledge all patient partners, research and clinical staff as well as the youth and families who have made this project possible.

\small
\bibliography{aaai22}

\end{document}

%% file: abstract.tex
We present an ongoing project that aims to develop a social robot to help children cope with painful and distressing medical procedures in a clinical setting. Our approach uses automated planning as a core component for action selection in order to generate plans that include physical, sensory, and social actions for the robot to use when interacting with humans. 
A key capability of our system is that the robot's behaviour adapts based on the affective state of the child patient.
The robot must operate in a challenging physical and social environment where appropriate and safe 
interaction with children, parents/caregivers, and healthcare professionals is crucial. 
In this paper, we present our system, examine some of the key challenges of the scenario, and describe how they are addressed by our system.

%% file: introduction.tex
Children regularly experience pain and distress in clinical settings, which can produce negative effects in both the short term (e.g., fear, distress, inability to perform procedures) and the long term (e.g., needle phobia, anxiety) \cite{stevens2011epidemiology}. While a range of techniques have been shown to help manage such situations (e.g., breathing exercises, distraction techniques, cognitive-behavioural interactions~\cite{chambers2009psychological}), delivered through a variety of means (e.g., distraction cards, kaleidoscopes, music, and virtual reality games), recent studies have also demonstrated that social robots can be used to manage child pain and distress during medical procedures~\cite{ali2019lo63,trost2019socially}.

We are developing a social robot to help children cope with painful and distressing medical procedures in a clinical setting \cite{10.1007/978-3-030-62056-1_45}.
The scenario presents a significant challenge for a social robot: the system must coexist
with multiple humans engaged in numerous high-priority and dynamic tasks. The robot behaviour
must be sensitive to the situation, as inappropriate behaviour may impact patient safety 
and well-being. Sensing the social state also presents a challenge:
not only are there multiple people, many likely wearing facial coverings, but the physical
space and processing bandwidth are also likely to be constrained. 
This situation is compounded by the fact that it may not always be clear how a child might react in a situation. 


To address these challenges, we underpin the robot's behaviour with an automated planning system that uses observed social signals, together with the robot's state, to select appropriate behaviour: the planner makes high-level decisions as to which spoken, non-verbal, and task-based actions should be taken next by the system. A key aspect of our approach is that the planner makes action selections not only based on the state of the world, but also using its beliefs about the developing interaction, as well as observations of the patient's affective state. 
The sensing components of the system are also designed to work in the target context,
ensuring the best possible input to the planning system.

This paper presents our ongoing work developing this robot system.
We give an overview of the target scenario and the system, which includes sensors, social signal processing, a web-based GUI, a planning system and a NAO robot.
We identify the main challenges that we have faced in designing the system for the clinical setting, including a lack of clear \emph{interaction landmarks} (important factors for structuring the interaction), the need for robustness, the challenges for learning predictive models, and the difficulties of integrating social signals into the planning model. 


%% file: related_work.tex
Technological systems based on Socially Assistive Robotics (SAR) \cite{Feil-Seifer.Mataric:2005} provide unique opportunities to establish new mechanisms that use human-like social communication as a means to generate embodied interaction. This type of Human-Robot Interaction (HRI) is considered potentially useful to create a shared relationship without touching the human, by using characteristics such as expressiveness, personality, dialogue, empathy and adaptation skills. Although it is not well established which particular elements of HRI dynamics produce changes in human behaviour, there are several studies that have reported benefits in various domains, such as social, behavioural, physical, and cognitive well-being in different populations~\cite{Amirova2021, Henschel2021}, in applications such as robot-assisted education~\cite{Johal2020}, autism diagnosis and therapy~\cite{Scassellati2018,GomezEsteban2017,Pennisi2016}, and Alzheimer therapy and elderly care~\cite{Tapus.etal:2009,Wada.etal:2004}.

Our work aims to enable the use of SAR in paediatric healthcare settings to help alleviate children's distress and pain. Despite the potential benefits of such an approach,
there are few studies in this area \cite{ali2019lo63,Jibb.etal:2018}, and almost none that use AI techniques to select the robot behaviour. \citet{trost2019socially} reported a review that included eight studies where a robot was used in this context: while the results seem promising and suggest that the robots succeeded in reducing pain, a need for improved methodology and measures was identified. 
Subsequently, \citet{trost2020socially} report on a study applying an empathic robot in real-world settings in an attempt to reduce paediatric pain and distress related to medical procedures. As results, the authors reported no significant difference on the mean scores of pain and distress scales between the study groups (distractive SAR vs empathic SAR). However, the authors suggest that empathic SAR could be clinically more effective since a greater willingness of children to the procedure was observed in this condition.

On the technical side, the idea of using planning to support interaction has a long history, and planning techniques have been applied previously in a range of social robots and interactive systems. Recent examples include~\cite{waldhart2016novel,Sanelli:2017,kominis2017multiagent,papaioannou2018human}. The most similar approach to ours is the JAMES social robot bartender~\cite{petrick2013planning,petrick2020knowledge}, which directly used an automated planner to choose the robot’s physical, sensing, and interactive actions. This system will form the basis of the approach used on this project. Recent work on explainable planning~\cite{fox2017explainable} has also highlighted the links between planning and user interaction, and is relevant to this work

%% file: system_overview.tex
In the specific clinical scenarios that we are targeting, the robot is
placed in a small room together with the patient (Figure~\ref{fig:scenario}, left), along with one or more carers and
healthcare providers, during the course of a single clinical procedure such as IV
insertion. Within this context, the robot must be able to adapt to different
roles throughout the intervention. Initially, the robot could behave as a mediator, introducing or explaining parts of a procedure. In another stage, the robot could behave as an assistant, performing actions alongside humans. The robot could also act as a tutor/interviewer at the end of a procedure in a debrief phase.


Our system architecture (see Figure \ref{fig:system}) is composed of several components, including social signal processing, an interaction manager, a planning system, and a robot platform. The target robot platform is the SoftBank NAO, which is a humanoid robot with 25 degrees of freedom, 
which enables it to move and perform a large variety of actions. Additionally, NAO is equipped with a speaker, allowing the generation of different stimuli using multiple communication channels, for example, using verbal language such as speech and body language through gestures. 

The low-level face analysis behaviour module is responsible for detecting the patient's face, identifying facial landmarks, head pose, gaze direction, and facial expression. Based on the above facial features, the social signal processing module estimates the current focus of attention and the head movement speed. This information is used to estimate the patient's emotional state, providing an indirect measure of affective states such as anxiety, valence, arousal, and engagement which are needed to control system behaviour. 

The estimation of social signals is highly uncertain, meaning that this type of signal can be ambiguous. A web-based application (Figure~\ref{fig:gui}) has therefore been implemented to provide an alternate input module that allows a research assistant or healthcare provider to generate and/or confirm the predictions. 
For example, the interface can be used to input the state of patient anxiety, or to pop up a window asking the user to confirm the completion of a clinical step.
The manual GUI-based module and the automated sensor-based module work simultaneously and complement each other to define the states needed for the decision-making process.

At the centre of the architecture is the interaction manager, which ensures synchronised transitions between the internal states of the system/robot. 
The interaction manager integrates the information from the social signal components to estimate the affective state. 
It also makes requests of the planning module, which is used during the interaction to determine the next action based on the current state and the goal. 
Finally, the social stimuli module interprets high-level actions and generates specific signals for each communication channel, whether through synthesised speech or non-verbal communication through gestures and body language.

The components have been implemented using embedded hardware to increase the processing capabilities of the NAO while maintaining the portability and flexibility of this platform. Additionally, an external RGB-D camera has been incorporated to complement the NAO's limited internal cameras (Figure~\ref{fig:scenario}, right). The framework was implemented using the Robot Operating System (ROS) \cite{Quigley:2009}, an open source standard middleware well known in the robotics community for its flexibility and scalability. 

%% file: procedure_modelling.tex

\begin{figure}[t]
    \centering
    \includegraphics[height=4cm]{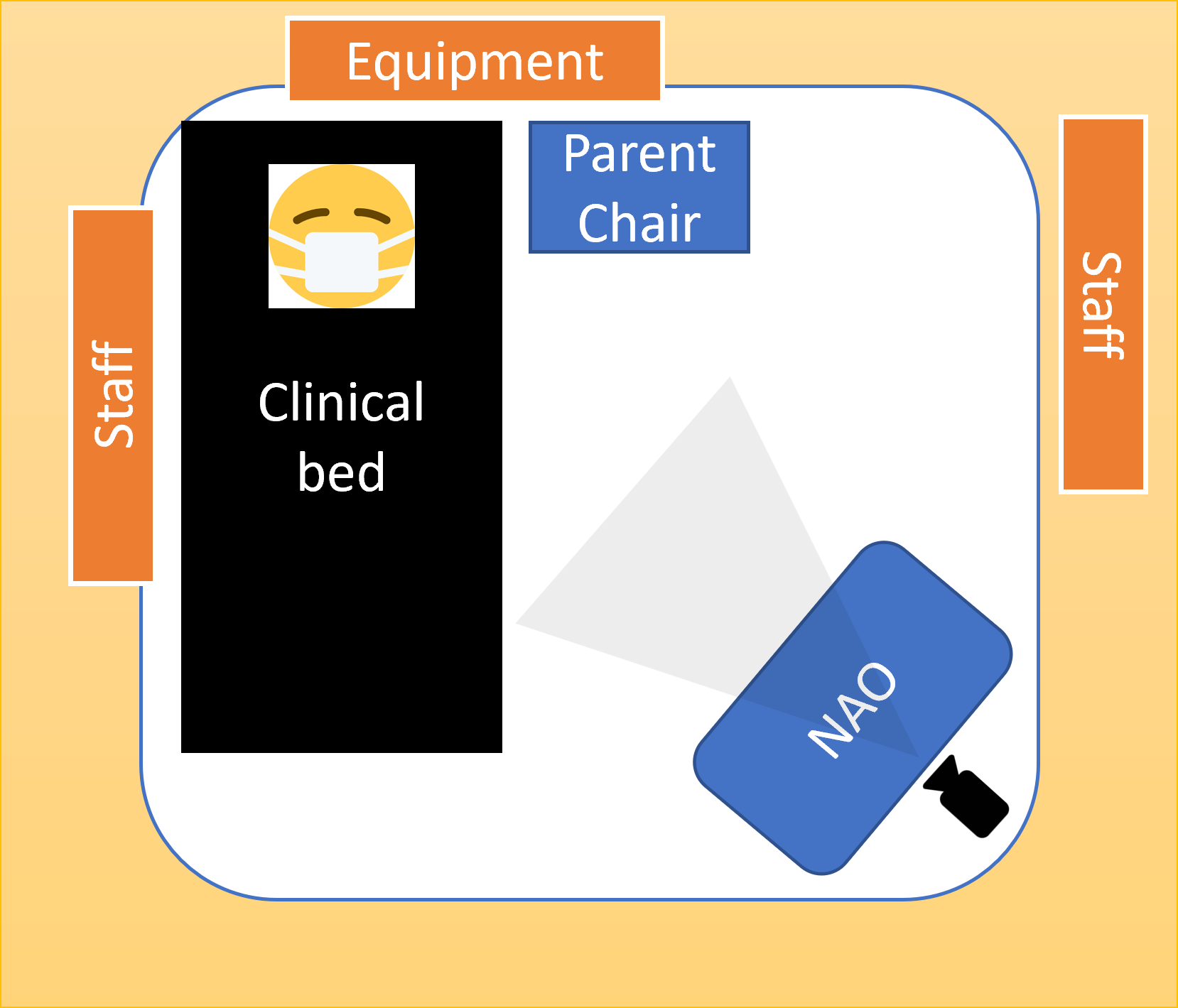}
    \includegraphics[height=4cm]{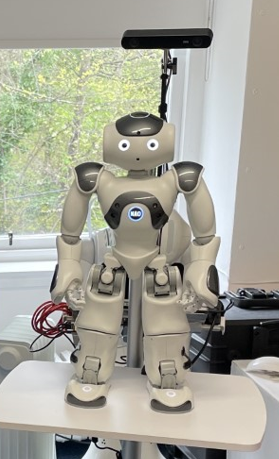}
    \caption{Target physical setting}
    \label{fig:scenario}
\end{figure}

\subsubsection{Interaction Modelling}

\begin{figure}[t]
    \centering
    \includegraphics[width=0.47\textwidth]{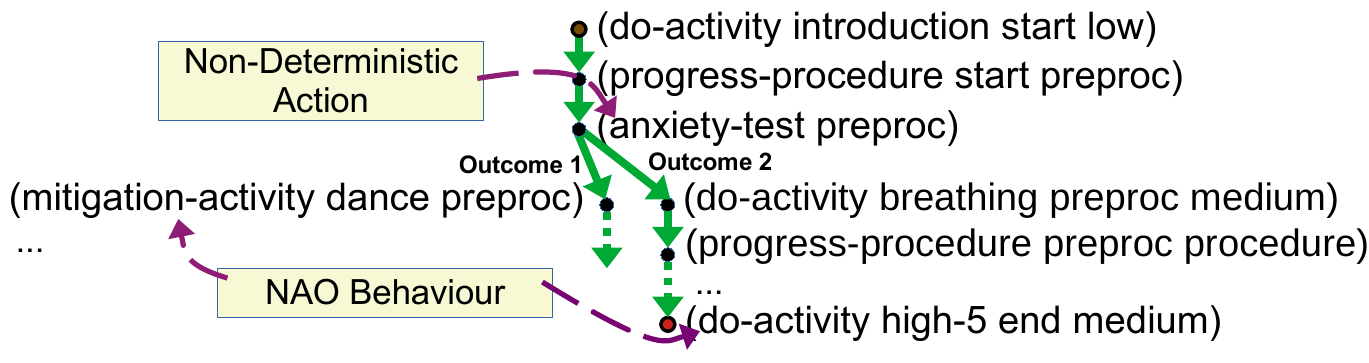}
    \caption{A partial plan showing NAO behaviour actions (e.g., breathing exercises and high five), non-deterministic actions (e.g., testing patient anxiety) and procedure actions.}
    \label{fig:plan}
\end{figure}

The robot's behaviours are underpinned by a planning model which uses a declarative representation to concisely represent the domain knowledge and possible interactions.
We use a fully observable non-deterministic planning model based on
\cite{muise2012improved}, which
can be defined as a tuple $\langle \mathcal{F},\mathcal{I}, \mathcal{G}, \mathcal{A}\rangle$, with fluents $\mathcal{F}$, initial state $\mathcal{I}$ (a full assignment to $\mathcal{F}$), a partial goal state $\mathcal{G}$, and a set of actions $\mathcal{A}$. Each action $a\in \mathcal{A}$ is a pair $\langle \mathit{pre}_a, \mathit{eff}_a\rangle$, with a precondition $\mathit{pre}_a$ (a subset of $\mathcal{F}$ that must hold) and an effect $\mathit{eff}_a$ (a set of possible outcomes---fluents that are made true or false).
If an action defines one outcome it is a deterministic action (see Figure~\ref{fig:action}); otherwise, it is a non-deterministic action.
Each action application results in an outcome, but the outcome cannot be chosen by the planner. 
A solution to the problem is instead a branched plan $\pi$, which includes alternative action outcomes and describes the sequence of actions that will achieve the goal, given any outcome (see Figure~\ref{fig:plan}). 

\begin{figure}[t]
\begin{Verbatim}[fontsize=\scriptsize,commandchars=\\\{\}]
(\textbf{:action} do-activity
  \textbf{:parameters} (?a - activity ?p - procstep ?x - level)
  \textbf{:precondition} (\textbf{and}
    (\textbf{not} (done ?a)) (procstage ?p) (desiredstrength ?p ?x)
    (okanxiety ?p) (naustep) (distractionstrength ?a ?x))
  \textbf{:effect} (\textbf{and}
    (\textbf{not} (naustep)) (done ?a) (procedurestep)))
\end{Verbatim}
\caption{A deterministic PDDL action representing a behaviour to be performed by the robot.}
\label{fig:action}
\end{figure}

Fluents model the situation in the room and state of the procedure (e.g., a health care provider is in the room), abstract information (e.g., that a certain behaviour has already been used), and affective state (e.g., anxiety or engagement of the patient).
Actions can be separated into four groups: robot behaviour, procedure update, implicit signals and explicit queries. 
The robot can perform a range of actions including distracting actions (e.g., dancing) and  calming and instructive actions (e.g., stepping through breathing exercises), each of which is represented in the planning model.

The system also has actions that can measure the affective state, 
which allows the system to identify issues in the interaction (e.g. a patient is experiencing higher than expected anxiety). This in turn allows it to choose mitigating actions (e.g., attempting to distract with a highly distracting action) and to monitor and ensure that the interception made the intended impact (e.g., reducing the anxiety of the patient).

%% file: world_knowledge.tex
\subsubsection{World State Estimation}

The world state representation is composed of predicted social cues associated with the patient's emotional state, engagement, and  willingness to participate in the procedure. However, the inherent complexity of the
target clinical procedure and uncertainty of the predictions require special care to ensure the correct interpretation of the scenario. As a result, a support tool is required to acquire additional information from the world when it is not possible to clearly estimate the predictions due to uncertainty; the interface of this tool is shown in Figure~\ref{fig:gui}.

Automated prediction of the patient state in this setting is challenging mainly due to occlusion: the patient is likely to be wearing a surgical mask, and there may be a constant flow of staff in the room. Additionally, the use of the space close to the patient is limited, which means that the equipment (e.g., the camera) must be located at some distance. Internet use is also limited by interference generated by the high flow of devices using the wireless network in the room.

With these limitations in mind, a pipeline was developed to automatically analyse the patient's face, along with a web interface to interact directly with the user, to confirm the estimated state. The automatic analysis pipeline is based on Nvidia DeepStream SDK\footnote{\url{https://developer.nvidia.com/deepstream-sdk}} and was deployed using a Jetson Nano board. In practice, six facial expressions, the focus of visual attention, and the speed of movement of the patient's head are estimated. From these estimations, characteristics such as engagement and state of anxiety are defined.



%% file: transition_modelling.tex
\subsubsection{Transition Modelling}
A particular challenge in this scenario has been the definition of the planning model. This is because it is difficult to define \emph{interaction landmarks}, which are model elements (e.g., the fluents $\mathcal{F}$) that can be clearly used to structure the interaction. For example, specifying in advance how the patient's affective state might be usefully exploited in the generation of appropriate interactions requires input from domain experts, who are not necessarily able to express the necessary knowledge in a way that can easily be represented formally for use in decision making. 

The system has therefore been designed to allow parameterisation, so that elements of the planning model can be associated with alternative transition implementors. In the current system, the options are: modelled, GUI provided, and sensed. These labels determine how each element is updated when an action is applied. Modelled elements are updated using information from the planning model; the values of world-determined (sensed and GUI provided) elements are gathered after the action has been applied, either by a predictive model or querying the user. 

Updates are performed by first identifying the most appropriate outcome. World elements are determined first, with the effects used to select the outcome that is most consistent with the observed effects. The planning model is then used to update the modelled elements. A key benefit of this approach is that it has enabled us to use the system in the early stages of development to demonstrate aspects of the proposed solution. In these demonstrations, we configured the system so that most elements were defined using the planning model. We could then investigate how the addition of certain information (e.g., child anxiety) could be incorporated into the model and used to influence the interactions. 

Such initial demonstrations have allowed us to receive quality feedback from key stakeholders through a co-design process, where discussions can be based on concrete and realistic system demonstrations. It is clear from these stakeholder discussions that patient anxiety is a crucial factor for action selection. At the moment, the system is configured to query the user for this information, while we are currently in the process of investigating a predictive model.

%% file: robustness.tex
\subsubsection{Robustness}

The eventual goal of this project is to evaluate the final robot system in a clinical trial, so robustness is crucial for appropriate behaviour: the robot will be used in real patient procedures, where the outcome of the robot performing inappropriate behaviour may impact patient safety and well-being. Moreover, technical difficulties can lead to challenges in interpreting the trial's results.

At the technical level, we have designed the system to use a series of connected components. This separation enables the clear allocation of responsibilities to individual components of the system, and allows us to test each component and error check the messages being passed between them. However, it also makes the interfaces between components vulnerable. For example, the interaction manager (see Figure~\ref{fig:system}) requires input from the GUI or sensors, and must allow time for a response. In practice, each of these  components may become unresponsive and fail to respond, for example due to network issues or robot failures.

As mitigation for these issues, we have incorporated explicit timeouts, default behaviours, and synchronisation. Each action type is associated with an explicit timeout and a default behaviour. After the time has elapsed, the system checks whether the appropriate message has been received and if not, the default behaviour is applied. The default behaviour will typically create an appropriate message type, populating the message fields using contextual information, including the parameters of the action. 
The interaction manager also plays a key role in coordinating the functions of the components and must also ensure that its internal state remains consistent. We have used keys and critical sections in the manager's code to ensure that the various threads are synchronised: for example, to ensure that a single output response is generated for every turn, and that a default does not interfere with specified GUI or robot behaviour.
The design of the system also provides clear and direct ways to stop the robot at any time in the interaction, if necessary.

An additional issue that is currently being investigated is how to best respond when the world facts do not correspond with an expected outcome. Although this problem is rather universal when using planning in robot control, it is often possible to use a simple process to combine previous experience with current observations.
In this case, the main issue lies in the fact that several of the elements may be modelled and used by the planner (i.e., they represent abstract states that cannot be observed). This means that the state at any given point must be combined from information from the world and the modelled parts. Our approach is to use the observations from the world (GUI and sensed) to estimate the appropriate impact on the modelled part of the state. In particular, if a change is made in the world that falls outside the expected boundaries, the system uses a series of rules to ensure that the modelled part of the state is consistent. 

%% file: conclusion.tex
\section{Conclusion}
This paper describes ongoing work aimed at developing a social robot to help children cope with painful and distressing medical procedures.
The scenario combines a dynamic and uncertain environment, complex social interaction that is difficult to specify fully in advance, and a real-world deployment location where robust and appropriate behaviour is crucial at every level.
We have described how our system design addresses these challenges: incorporating social signals into the planning model, providing a web-based GUI to provide an alternative to sensing the world and the patient's social signals, configuring elements of the approach to allow for the easy selection of an information source (real observations or modelled), and by using a targeted strategy to address the robustness of the system.
The first version of the system is currently undergoing usability testing as we continue to develop the system components, interaction model, and appropriate predictive models for social signals. When the system is complete, we plan to test its feasibility in a two-site clinical trial in paediatric emergency departments.

%% file: AI-HRI_2022.bbl
\begin{thebibliography}{25}
\providecommand{\natexlab}[1]{#1}

\bibitem[{Ali et~al.(2019)Ali, Manaloor, Ma, Sivakumar, Vandermeer, Beran,
  Scott, Graham, Curtis, Jou et~al.}]{ali2019lo63}
Ali, S.; Manaloor, R.; Ma, K.; Sivakumar, M.; Vandermeer, B.; Beran, T.; Scott,
  S.; Graham, T.; Curtis, S.; Jou, H.; et~al. 2019.
\newblock LO63: humanoid robot-based distraction to reduce pain and distress
  during venipuncture in the pediatric emergency department: a randomized
  controlled trial.
\newblock \emph{Canadian Journal of Emergency Medicine}, 21(S1): S30--S31.

\bibitem[{Amirova et~al.(2021)Amirova, Rakhymbayeva, Yadollahi, Sandygulova,
  and Johal}]{Amirova2021}
Amirova, A.; Rakhymbayeva, N.; Yadollahi, E.; Sandygulova, A.; and Johal, W.
  2021.
\newblock 10 Years of Human-NAO Interaction Research: A Scoping Review.
\newblock \emph{Frontiers in Robotics and AI}, 8.

\bibitem[{Chambers et~al.(2009)Chambers, Taddio, Uman, McMurtry, and
  {HELPinKIDS Team}}]{chambers2009psychological}
Chambers, C.~T.; Taddio, A.; Uman, L.~S.; McMurtry, C.~M.; and {HELPinKIDS
  Team}. 2009.
\newblock Psychological interventions for reducing pain and distress during
  routine childhood immunizations: a systematic review.
\newblock \emph{Clinical therapeutics}, 31: S77--S103.

\bibitem[{Feil-Seifer and Mataric(2005)}]{Feil-Seifer.Mataric:2005}
Feil-Seifer, D.; and Mataric, M. 2005.
\newblock Socially Assistive Robotics.
\newblock In \emph{9th International Conference on Rehabilitation Robotics,
  2005. ICORR 2005.} IEEE.

\bibitem[{Foster et~al.(2020)Foster, Ali, Litwin, Parker, Petrick, Smith,
  Stinson, and Zeller}]{10.1007/978-3-030-62056-1_45}
Foster, M.~E.; Ali, S.; Litwin, S.; Parker, J.; Petrick, R. P.~A.; Smith,
  D.~H.; Stinson, J.; and Zeller, F. 2020.
\newblock Using AI-Enhanced Social Robots to Improve Children's Healthcare
  Experiences.
\newblock In \emph{Social Robotics}, 542--553.

\bibitem[{Fox, Long, and Magazzeni(2017)}]{fox2017explainable}
Fox, M.; Long, D.; and Magazzeni, D. 2017.
\newblock Explainable planning.
\newblock In \emph{IJCAI Workshop on XAI}.

\bibitem[{Gomez~Esteban et~al.(2017)Gomez~Esteban, Baxter, Belpaeme, Billing,
  Cai, Cao, Coeckelbergh, Costescu, David, De~Beir, Fang, Ju, Kennedy
  et~al.}]{GomezEsteban2017}
Gomez~Esteban, P.; Baxter, P.~E.; Belpaeme, T.; Billing, E.; Cai, H.; Cao,
  H.-L.; Coeckelbergh, M.; Costescu, C.; David, D.; De~Beir, A.; Fang, Y.; Ju,
  Z.; Kennedy, J.; et~al. 2017.
\newblock {How to Build a Supervised Autonomous System for Robot-Enhanced
  Therapy for Children with Autism Spectrum Disorder}.
\newblock \emph{Paladyn Journal of Behavioral Robotics}, 8(1): 18--38.

\bibitem[{Henschel, Laban, and Cross(2021)}]{Henschel2021}
Henschel, A.; Laban, G.; and Cross, E.~S. 2021.
\newblock What Makes a Robot Social? A Review of Social Robots from Science
  Fiction to a Home or Hospital Near You.
\newblock \emph{Current Robotics Reports}, 2: 9--19.

\bibitem[{Jibb et~al.(2018)Jibb, Birnie, Nathan, Beran, Hum, Victor, and
  Stinson}]{Jibb.etal:2018}
Jibb, L.~A.; Birnie, K.~A.; Nathan, P.~C.; Beran, T.~N.; Hum, V.; Victor,
  J.~C.; and Stinson, J.~N. 2018.
\newblock Using the {MEDiPORT} humanoid robot to reduce procedural pain and
  distress in children with cancer: A pilot randomized controlled trial.
\newblock \emph{Pediatric Blood \& Cancer}, 65(9): e27242.

\bibitem[{Johal(2020)}]{Johal2020}
Johal, W. 2020.
\newblock Research Trends in Social Robots for Learning.
\newblock \emph{Current Robotics Reports}, 1: 75--83.

\bibitem[{Kominis and Geffner(2017)}]{kominis2017multiagent}
Kominis, F.; and Geffner, H. 2017.
\newblock Multiagent online planning with nested beliefs and dialogue.
\newblock In \emph{Twenty-Seventh International Conference on Automated
  Planning and Scheduling}.

\bibitem[{Muise, McIlraith, and Beck(2012)}]{muise2012improved}
Muise, C.; McIlraith, S.; and Beck, C. 2012.
\newblock Improved non-deterministic planning by exploiting state relevance.
\newblock In \emph{Proceedings of the International Conference on Automated
  Planning and Scheduling}, 172--180.

\bibitem[{Papaioannou, Dondrup, and Lemon(2018)}]{papaioannou2018human}
Papaioannou, I.; Dondrup, C.; and Lemon, O. 2018.
\newblock Human-robot interaction requires more than slot
  filling-multi-threaded dialogue for collaborative tasks and social
  conversation.
\newblock In \emph{FAIM/ISCA Workshop on Artificial Intelligence for Multimodal
  Human Robot Interaction}, 61--64.

\bibitem[{Pennisi et~al.(2016)Pennisi, Tonacci, Tartarisco, Billeci, Ruta,
  Gangemi, and Pioggia}]{Pennisi2016}
Pennisi, P.; Tonacci, A.; Tartarisco, G.; Billeci, L.; Ruta, L.; Gangemi, S.;
  and Pioggia, G. 2016.
\newblock {Autism and social robotics: A systematic review}.
\newblock \emph{Autism Research}, 9(2): 165--183.

\bibitem[{Petrick and Foster(2020)}]{petrick2020knowledge}
Petrick, R.; and Foster, M.~E. 2020.
\newblock Knowledge engineering and planning for social human--robot
  interaction: A case study.
\newblock In \emph{Knowledge Engineering Tools and Techniques for AI Planning},
  261--277. Springer.

\bibitem[{Petrick and Foster(2013)}]{petrick2013planning}
Petrick, R.~P.; and Foster, M.~E. 2013.
\newblock Planning for Social Interaction in a Robot Bartender Domain.
\newblock In \emph{Proceedings of the International Conference on Automated
  Planning and Scheduling}.

\bibitem[{Quigley et~al.(2009)Quigley, Conley, Gerkey, Faust, Foote, Leibs,
  Wheeler, and Ng}]{Quigley:2009}
Quigley, M.; Conley, K.; Gerkey, B.; Faust, J.; Foote, T.; Leibs, J.; Wheeler,
  R.; and Ng, A.~Y. 2009.
\newblock {ROS}: an open-source {R}obot {O}perating {S}ystem.
\newblock In \emph{{ICRA} Workshop on Open Source Software}.

\bibitem[{Sanelli et~al.(2017)Sanelli, Cashmore, Magazzeni, and
  Iocchi}]{Sanelli:2017}
Sanelli, V.; Cashmore, M.; Magazzeni, D.; and Iocchi, L. 2017.
\newblock Short-Term Human-Robot Interaction through Conditional Planning and
  Execution.
\newblock In \emph{Proceedings of the International Conference on Automated
  Planning and Scheduling (ICAPS)}, 540--548.

\bibitem[{Scassellati et~al.(2018)Scassellati, Boccanfuso, Huang, Mademtzi,
  Qin, Salomons, Ventola, and Shic}]{Scassellati2018}
Scassellati, B.; Boccanfuso, L.; Huang, C.-M.; Mademtzi, M.; Qin, M.; Salomons,
  N.; Ventola, P.; and Shic, F. 2018.
\newblock {Improving social skills in children with ASD using a long-term,
  in-home social robot}.
\newblock \emph{Science Robotics}, 3(21): eaat7544.

\bibitem[{Stevens et~al.(2011)Stevens, Abbott, Yamada, Harrison, Stinson,
  Taddio, Barwick, Latimer, Scott, Rashotte et~al.}]{stevens2011epidemiology}
Stevens, B.~J.; Abbott, L.~K.; Yamada, J.; Harrison, D.; Stinson, J.; Taddio,
  A.; Barwick, M.; Latimer, M.; Scott, S.~D.; Rashotte, J.; et~al. 2011.
\newblock Epidemiology and management of painful procedures in children in
  Canadian hospitals.
\newblock \emph{Cmaj}, 183(7): E403--E410.

\bibitem[{Tapus, Tapus, and Mataric(2009)}]{Tapus.etal:2009}
Tapus, A.; Tapus, C.; and Mataric, M.~J. 2009.
\newblock The use of socially assistive robots in the design of intelligent
  cognitive therapies for people with dementia.
\newblock In \emph{2009 IEEE International Conference on Rehabilitation
  Robotics}. IEEE.

\bibitem[{Trost et~al.(2020)Trost, Chrysilla, Gold, and
  Matari{\'c}}]{trost2020socially}
Trost, M.~J.; Chrysilla, G.; Gold, J.~I.; and Matari{\'c}, M. 2020.
\newblock Socially-Assistive Robots Using Empathy to Reduce Pain and Distress
  during Peripheral IV Placement in Children.
\newblock \emph{Pain research \& management}, 1--7.

\bibitem[{Trost et~al.(2019)Trost, Ford, Kysh, Gold, and
  Matari{\'c}}]{trost2019socially}
Trost, M.~J.; Ford, A.~R.; Kysh, L.; Gold, J.~I.; and Matari{\'c}, M. 2019.
\newblock Socially assistive robots for helping pediatric distress and pain: a
  review of current evidence and recommendations for future research and
  practice.
\newblock \emph{The Clinical journal of pain}, 35(5): 451.

\bibitem[{Wada et~al.(2004)Wada, Shibata, Saito, and Tanie}]{Wada.etal:2004}
Wada, K.; Shibata, T.; Saito, T.; and Tanie, K. 2004.
\newblock Effects of robot-assisted activity for elderly people and nurses at a
  day service center.
\newblock \emph{Proceedings of the IEEE}, 92(11): 1780–1788.

\bibitem[{Waldhart, Gharbi, and Alami(2016)}]{waldhart2016novel}
Waldhart, J.; Gharbi, M.; and Alami, R. 2016.
\newblock A novel software combining task and motion planning for human-robot
  interaction.
\newblock In \emph{2016 AAAI Fall Symposium Series}.

\end{thebibliography}
